\pdfoutput=1
\documentclass[sigconf]{acmart}

\usepackage[utf8]{inputenc} % allow utf-8 input
\usepackage[T1]{fontenc}    % use 8-bit T1 fonts
\usepackage{hyperref}       % hyperlinks
\usepackage{url}            % simple URL typesetting
\usepackage{amsmath}
\usepackage{booktabs}       % professional-quality tables
\usepackage{amsfonts}       % blackboard math symbols
\usepackage{nicefrac}       % compact symbols for 1/2, etc.
\usepackage{microtype}      % microtypography
\usepackage{rotating,graphicx,bbm,algorithmic,algorithm}
\usepackage{amsthm,amstext,amsfonts,amscd,color,ulem}
\usepackage{booktabs,ragged2e,caption,footnote,multirow} %ltablex
\usepackage{subcaption}
\usepackage{tcolorbox}
\usepackage{balance}

\newtheorem{theorem}{Theorem}[section]
\newtheorem{lemma}[theorem]{Lemma}

\definecolor{dgreen}{rgb}{0,0.5,0}

\makeatletter
\newcommand{\vast}{\bBigg@{4}}
\newcommand{\Vast}{\bBigg@{5}}

\newcommand{\eat}[1]{}

\newcommand{\minimize}[1]{\underset{#1}{\mbox{minimize}}}

\newcommand{\argmin}[1]{\underset{#1}{\mbox{argmin}}}

\AtBeginDocument{%
  \providecommand\BibTeX{{%
    \normalfont B\kern-0.5em{\scshape i\kern-0.25em b}\kern-0.8em\TeX}}}

\setcopyright{rightsretained}
\begin{document}

\title{Lambda Learner: Fast Incremental Learning on Data Streams}


\author{ Rohan Ramanath, Konstantin Salomatin, Jeffrey D. Gee, Kirill Talanine}
\author{Onkar Dalal, Gungor Polatkan, Sara Smoot, Deepak Kumar}
\email{ {rramanath,ksalomatin,jgee,odalal,gpolatkan,ssmoot,dekumar}@linkedin.com}

\affiliation{\vspace{0.1cm}
    \institution{LinkedIn Corporation, USA}
    }

\fancyhead{}

\renewcommand{\shortauthors}{Ramanath, et al.}

\begin{abstract}
One of the most well-established applications of machine learning is in deciding what content to show website visitors. When observation data comes from high-velocity, user-generated data streams, machine learning methods perform a balancing act between model complexity, training time, and computational costs. Furthermore, when model freshness is critical, the training of models becomes time-constrained. Parallelized batch offline training, although horizontally scalable, is often not time-considerate or cost-effective. In this paper, we propose Lambda Learner, a new framework for training models by incremental updates in response to mini-batches from data streams. We show that the resulting model of our framework closely estimates a periodically updated model trained on offline data and outperforms it when model updates are time-sensitive. We provide theoretical proof that the incremental learning updates improve the loss-function over a stale batch model. We present a large-scale deployment on the sponsored content platform for a large social network, serving hundreds of millions of users across different channels (e.g., desktop, mobile). We address challenges and complexities from both algorithms and infrastructure perspectives, illustrate the system details for computation, storage, stream processing training data and open-source the system.
\end{abstract}
%For example, when recommending content to users, it has been shown that personalization through random-effect models (i.e., per-member, per-content personalization) can be effective in real-world scenarios. However, random-effect models trained on large datasets can grow incredibly complex when personalizing over fast-growing member and content pools.
\begin{CCSXML}
<ccs2012>
   <concept>
       <concept_id>10003752.10003753.10003760</concept_id>
       <concept_desc>Theory of computation~Streaming models</concept_desc>
       <concept_significance>300</concept_significance>
       </concept>
   <concept>
       <concept_id>10002951.10003260.10003261.10003271</concept_id>
       <concept_desc>Information systems~Personalization</concept_desc>
       <concept_significance>100</concept_significance>
       </concept>
   <concept>
       <concept_id>10010147.10010257.10010282.10010284</concept_id>
       <concept_desc>Computing methodologies~Online learning settings</concept_desc>
       <concept_significance>500</concept_significance>
       </concept>
 </ccs2012>
\end{CCSXML}

\ccsdesc[300]{Theory of computation~Streaming models}
\ccsdesc[100]{Information systems~Personalization}
\ccsdesc[500]{Computing methodologies~Online learning settings}

\keywords{Mixed-effect Models; Incremental Online Learning; Recommender Systems; Personalization; Reinforcement Learning}

\copyrightyear{2021}
\acmYear{2021}
\acmConference[KDD '21]{Proceedings of the 27th ACM SIGKDD Conference on Knowledge Discovery and Data Mining}{August 14--18, 2021}{Virtual Event, Singapore}
\acmBooktitle{Proceedings of the 27th ACM SIGKDD Conference on Knowledge Discovery and Data Mining (KDD '21), August 14--18, 2021, Virtual Event, Singapore}\acmDOI{10.1145/3447548.3467172}
\acmISBN{978-1-4503-8332-5/21/08}


\maketitle

\section{Introduction}
    Content in a professional social network is distributed across the diverse member experiences the site has to offer, from advertising and recruiting, to job and knowledge seeking. Each experience depends on a suite of machine learned models, which rank and curate content to create quality experiences, typically measured by engagement. Since quality experiences are intrinsically member-specific, personalized models are more performant in most settings. 

A common challenge for recommender systems is to achieve \textit{memorization} and \textit{generalization}. Most state-of-the-art models \cite{wang2017deep,cheng2016wide} address generalization via embedding based methods \cite{rendle2010factorization,covington2016deep} or other nonlinear models and capture memorization via interaction features, ids of the user/item or a crossing network. The memorization piece is important because it is difficult to learn effective low-dimensional representations for users and items when the underlying user-item matrix is sparse and high-rank. Since the features used in the memorization part of the model could be both transient and ever growing, being able to update this part of the model is critical to performance. For example, a new advertising campaign can reach thousands of impressions in a few minutes. Likewise, a job posting can reach millions of job seekers in a similarly short time window. In these cases, recommendation models need to learn from the information as quickly as possible. Within the context of computational advertising, we specifically observe the following challenges when estimating viewer engagement:

\begin{itemize}
    \item \textbf{Cold start}: Over the course of a day, a model needs to make decisions on advertisers and advertisements not observed in the training set. These can be brand-new advertisers and advertisements, or advertisers who have not marketed through the social network for extended periods of time. Under these scenarios, the memorization piece of the model must quickly adapt to new data. 
    \item \textbf{Data shifts}: Viewers’ most recent engagements are the best interest indicators, as interest can dramatically shift over short periods of time. For example, a viewer who might be heavily engaging with advertisements about cloud services on the site may be less-inclined to engage immediately after signing up with some of the cloud services.
\end{itemize}

Although the prior work \cite{wang2017deep,cheng2016wide,rendle2010factorization,covington2016deep} acknowledges the need to re-train the overall model frequently (\citet{cheng2016wide} recommends warm start based fine-tuning and empirically checking performance as a sanity check), there is no model agnostic solution with theoretical guarantees of performance improvement. One class of models that have been successful for personalized recommendations \& ranking is Generalized Additive Mixed Effect models (GAME)\citep{zhang2016glmix}.  This consists of a typically large fixed-effects model (\textit{generalization}) that is trained on the whole dataset to improve the model's performance on previously unseen user-item pairs, and a series of simpler linear random-effects models (\textit{memorization}) trained on data corresponding to each entity (e.g. user or article or ad) for more granular personalization. The fixed effects part of the model can be arbitrarily complex (e.g. wide \& deep, boosted trees, ensembles, or linear \cite{zhang2016glmix}\footnote{GLMix is an instantiation of GAME with a linear fixed-effect model}) depending on the targeted task. The entity specific random-effect models (detailed in Section \ref{GAME}) are a powerful way to personalize both user and item (content/ad) patterns.
% However, in cases where models are fit to incredibly large datasets, the additional model-complexity comes with large computational costs.

The most general solution—frequent offline model retraining and deployment—often proves insufficient for the following reasons: (1) to model newly introduced content or content with short-lived significance, computation time becomes a valuable commodity, and limiting it causes a computationally-constrained learning problem \citep{bottou2010large}, i.e. the computational constraints introduced to retrain the model disallow us from meeting self-imposed update frequencies; (2) the cold-start problem can cause long-term repercussions for the site, i.e. poor recurring recommendations may influence future engagement for new or returning dormant members. Additionally, the model performance slowly degrades over time, demonstrating the need to update the model continuously. 

In this paper, we propose a framework to incrementally train the memorization part of the model as a booster over the generalization part. 
% In the process we show that updating the model using the loss on the incoming data (i.e. fine-tuning loss) is inadequate, as errors accumulate with each update.
We incrementally update the memorization model between full batch offline updates of the generalization model to balance training performance against model stability. We show that for a scenario that suffers from the aforementioned cold-start, incremental training enables quicker adaptation to new and short-lived data, and reduces the rate of model degradation.
The key concept is to prepare mini-batches of data from an incoming stream of logged data to train an incremental update to the memorization model. We approximate the loss function on previously observed data by a local quadratic approximation around the previous optimal value and combine it with the loss function on the most recent mini-batch. This allows us to incrementally update the model without relying on all historical data yet do better than just relying on the new incoming data. This results in a regularized learning problem with a weighted penalty. In the Bayesian interpretation, the local quadratic approximation is equivalent to using the posterior distribution of the previous model as prior for training the update.

\subsection {Related Work}

Similar problems have been studied from various angles. \citet{Gepperth2016IncrementalLA} survey incremental learning under streaming data, giving an overview of popular approaches, foundations, and applications. \citet{Nonstationary} provide a survey on non-stationary scenarios in which modeling frameworks  fail to address evolving or drifting dynamics (e.g., seasonality or periodicity effects, the user behavioral changes or preferences). Bayesian, stochastic, tree-based, and instance-based learning approaches have been explored in the online scenario with varying degrees of success on both stationary and non-stationary data \citep{losing2018incremental}. For potentially non-stationary data, it has been shown that Stochastic Gradient Descent (SGD) on a linear model best balanced performance and growth in model complexity \citep{losing2018incremental}. \citet{bottou2010large} demonstrated that SGD in very large datasets out-performed more modern approaches due to the speed of convergence (to a predefined ``expected risk''). It was also found that second-order SGD (2SGD) in a single-pass, under convexity and metric regularity assumptions was able to approximate iterative, full-batch SGD \citep{bottou2010large}.

%Robotics and similar resource-constrained hardware systems are important application domains for incremental learning. For example, \citet{bioRobot} uses Markov decision processes (MDPs) for incremental self-learning in robots, whereas \citet{mobileRobot} presents an architecture in high-dimensional covariate spaces, demonstrating similar feasibility in mobile robots. \citet{camera} use probabilistic graphical models to model activity in a network of cameras through incremental learning of time-delayed dependencies.

From an optimization point of view, \citet{McMahan} investigates different strategies for online convex optimization and finds that several such algorithms are actually highly related, being distinguishable based on how they handle $L1$ regularization. \citet{localIncremental} focus on a local strategy to avoid repeated learning of all parameters. The proposed algorithm selects a local subset to run incremental learning. \citet{largeIncremental} study model performance degradation due to missing past data and recommends using a data bias elimination mechanism to make incremental learning methods more effective.

From the representation learning point of view, \citet{adaptiveDeep} study incremental learning for deep models, focusing on flexible model evolution and updating models such that previous knowledge is not forgotten. They propose using a hidden attention layer. \citet{contextualRecs} consider the problem from collaborative member behavior and propose an item clustering-based system. Similarly, \citet{dhurandhar2014efficient} approach the problem of incremental feature learning for GLMs where a snapshot of batch data is available.
%Recommendations are selected from content clusters to which members are assigned based on their engagement patterns.
Though such approaches might fit more static content libraries, they do not suit more dynamic environments, such as an advertising platform in which decent relevance accuracy should be accomplished quickly once content is added to the ecosystem.  Online learning for recommending content from large and dynamic pools is studied by \citet{fastOnline}. A factor model is proposed to fit content-specific factors quickly using online regression. Though the online learning portion has similarities to our work, the problem does not consider large co-variate space. Lastly, a family of reinforcement learning algorithms could be used for incrementally updating recommender systems (by changing the objective function), \citet{chen2019top} propose a model applied to the candidate generation phase. Our work is strictly focused on the ranking phase, assuming independence of the candidate selection generation methodology, and fixing the objective function used to rank the ads. 

\subsection {Our Contributions}
In this paper, our contributions are in four fold: 1) We propose a new framework for incremental learning for personalization at scale. We perform model adjustments over the most recent data observations and adapt to the latest trends, additionally training new models for content unobserved in the training set. 2) We provide theoretical guarantees for improvements in loss-function for the incrementally updated model over the stale batch model. 3) We conduct experiments showing that incrementally updated models in a single pass over a data stream can closely approximate full batch offline models. 4) Finally, we present a large-scale deployment and impact on the ads platform at a large social network, address infrastructure challenges, and open-source the system\footnote{https://github.com/linkedin/lambda-learner}.
%illustrate system details for computation, storage, and nearline data-generation.

\begin{figure*}
  \includegraphics[width=0.7\textwidth]{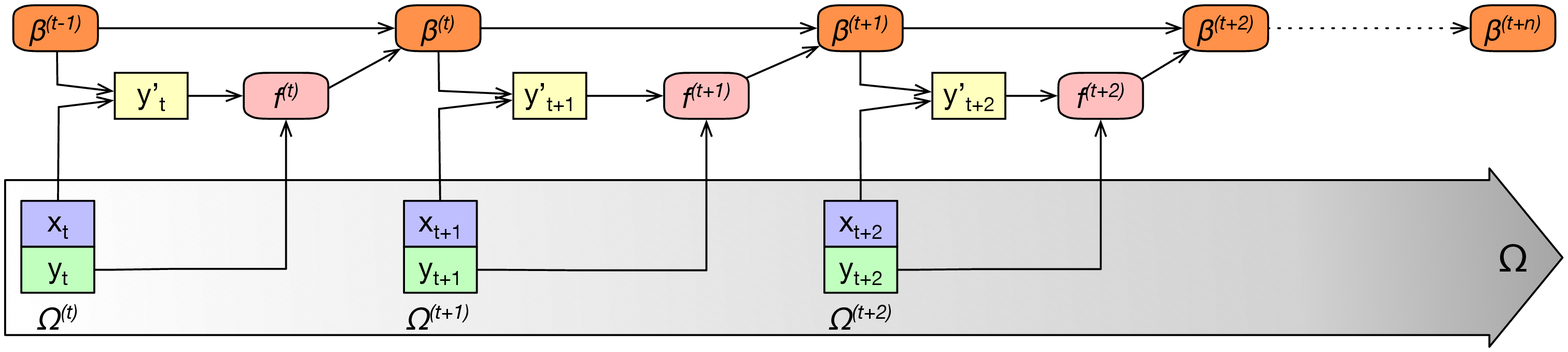}
  \vspace{-0.1in}
  \caption{An illustration of the incremental learning process}
  \label{fig:illustrate-incremental-learning}
  \vspace{+0.1in}
\end{figure*}

%\subsection {Organization}
%This paper is organized as follows. In Section \ref{Background}, we provide a high-level introduction to the foundation from which we derive our framework. In Section \ref{Lambda Learning}, we describe the Lambda Learner framework for incremental, single-pass, batch updates to random-effect models (described in Section \ref{GAME}). In Section \ref{Theory}, we provide the mathematical theory motivating the solution and prove convergence. In Section \ref{System Overview}, we describe our end-to-end system with application to ads-serving at a large social network. Lastly, in Section \ref{Experiments}, we show the results of offline experimentation under the algorithm proposed in Section \ref{Lambda Learning} on a public dataset, as well as a dataset sampled from the ad environment of a large social network.

\subsection {Terminology and Notation}
Throughout this document we will commonly reference the following notation. We refer to a training dataset $\Omega^{(t)}$, as the mini-batch/slice of ordered ``data instances'' observed in time interval $t$. Each $\Omega^{(t)}$ consists of data instances represented by the tuple $({\bf x_n}, y_n)$, where $\bf x_n$ is an observed feature vector describing instance $n$, and $y_n$ is a corresponding observed label. We denote the cumulative training dataset observed until time $t$ using $\bar{\Omega}^{(t)} = \bar{\Omega}^{(t-1)} \cup \Omega^{(t)}$. We use $\beta^{*(t)}$ to refer to a parameterized model trained on the full dataset $\bar{\Omega}^{(t)}$, and $\beta^{(t)}_{inc}$ to refer to the incremental model trained using the new Lambda Learner framework.

In the GAME framework, the model $\beta = [\beta_f, {\bf B}]$ consists of two components: a fixed-effects model, $\beta_f$, and a set of random-effect models, ${\bf B} = \{B_{r}\}_{r \in \mathcal{R}}$, where $\mathcal{R}$ denotes the set of random-effect types, see section~\ref{GAME} for details. Coefficients of a single random-effect type $r$ are denoted by $B_{r} = \{\beta_{rl}\}_{l=1}^{N_{r}}$, where $N_{r}$ is the total number of random-effects with random-effect type, $r$ (e.g. $r$ may correspond to ``viewers'', ``articles'', ``ads'' or ``jobs''; $l$ is a specific viewer, article, ad or job ID, and $N_r$ is the cardinality of viewers, articles, ads or jobs). Lastly, for each data instance ${{\bf x}_n, y_n}$, we use $i(r, n)$ to represent the index value of random-effect type $r$ for instance $n$.

\section{Background}\label{Background}
% Section 2
% Third, Try to write in a way that connects to the f, F notations used elsewhere 
% Fourth, the fixed and random effects stuff is not coming out clearly - You could explain it with an example, say the problem used in section 5 and the experiments. 
% "R" may be confused for the set of real numbers.
% To help the Intro (Section 1) you could even move this formulation to that section so that the reader will have clarity right from the beginning.
% I found figure 1 to be somewhat disconnected from the text. When you refer to the figure in Subsection 2.3, the reviewer's tendency would be to look at the figure immediately. But the notations there don't fit too well with what is in the text.
In this section, we review a high-level foundation of personalization through random effect models and incremental learning.
\subsection{Learning on Data Streams}
\label{test}
Given a stream of observation data $\Omega^{(0)}, \Omega^{(1)}, \ldots, \Omega^{(t)}, \ldots$, and a base convex loss function $f$ (e.g. log-logistic), we define a series of loss functions over the mini-batch of datastream $\Omega^{(t)}$ as
\begin{alignat}{1}
%\label{gen:main}
f_t(\beta) &=\sum_{n \in {\Omega}^{(t)}} f({\bf x_n}, y_n, \beta),
\end{alignat}
and a time-discounted cumulative loss over the total data $\bar{\Omega}^{(t)}$
\begin{alignat}{1}
%\label{gen:main-cumulative}
F_{t}(\beta) &= \sum_{k=0}^{k=t} \delta^{t-k} f_k({\beta}) \nonumber = \delta F_{t-1}(\beta) + f_t(\beta),
\end{alignat}
where, $\delta < 1$ is a time discount or forgetting factor to emphasize the fresher data.

\subsection{Generalized Additive Mixed Effect Models} \label{GAME}
\subsubsection{Generalized Linear Mixed Models}
Zhang et al \citep{zhang2016glmix} proposed Generalized Linear Mixed Models (GLMix) as an approach for personalizing traditional Generalized Linear Models (GLM's) through the addition of linear, ``random-effect" models, keyed to discretely valued random-effect variables. Example random-effect variables range from unique entity identifiers (e.g., advertisement, job, or user) to contextual information such as the consumption device (e.g., desktop, mobile application, etc). In practice, the coefficients fit conditionally on the values of these variables and serve a supplementary role in providing personalization atop the GLM.

We solve for the fixed-effect coefficients (unconditional part of the model), $\beta_f$, and the random-effect coefficients, ${\bf B}$, through the optimization of negative log likelihood objective
\begin{alignat}{1}
\label{glmix:main}
\minimize{\beta = [\beta_f, {\bf B}]} & \sum_{n \in \bar{\Omega}} - \log p(y_n | s_n) - \log p(\beta_f) - \sum_{r \in {\mathcal R}} \sum_{l=1}^{N_r} \log p(\beta_{rl})
\end{alignat}
where the first summation term is data likelihood and the score $s_n$ is a sum of fixed-effect and random-effect scores for the $n$-th observation:
\begin{alignat}{1}
\label{glmix:sub}
s_n &= {\bf x}_n^T\beta_f + \sum_{r \in {\mathcal R}} {\bf z}_{rn}^{T}\beta_{r,i(r,n)},
\end{alignat}
${\bf x}_{n}$ is the feature vector for the fixed-effect of observation $n$, $z_{rn}$ is the feature vector for random-effect type $r$ of $n$, and $\beta_{r,i(r,n)}$ is the coefficients vector conditional on the random-effect variable value $i(r,n)$. For example, in a user-item recommendation model with per-user and per-item random effects: $r\in\{"user","item"\}$ and $i("user",n)$ is simply the id of the user in the observation $n$, and $i("item",n)$ - id of the item. The latter terms terms $\log p(\beta_f)$ and $\log p(\beta_{rl})$ in~\eqref{glmix:main} represent the prior of the model coefficients, in particular, a Gaussian prior $\mathcal{N} (0, \lambda I)$.

\subsubsection{Generalized Additive Mixed Effect}
Although the GLMix formulations in (\ref{glmix:main}) and (\ref{glmix:sub}) describe fixed-effect and random-effects as linear models, they are not limited to such. By substituting $s_{n}$ for:
\begin{alignat}{1}
\label{glmix:generalsub}
s_n &= g_f(x_{n},\beta_f) + \sum_{r \in {\mathcal R}} g_r({\bf z}_{rn}, \beta_{r,i(r,n)}),
\end{alignat}
we define the Generalized Additive Mixed Effect (GAME) model class, where $g_f$ and $g_r$ are arbitrary scoring functions. For example, $g_f$ can be a deep neural network or a bag of decision trees. GLMix is a purely linear example of GAME.

\subsection{Incremental  Learning}
Incremental Learning (IL) describes the general set of algorithms attempting to solve (often) large-scale machine learning through immediate model updates in response to new data instances (Figure \ref{fig:illustrate-incremental-learning}), without additional passes over pre-existing data. These scenarios occur when training models on large streams of data, where there is a limited time window for models to adjust, constraining computational cost of the learning process. Model update rates can be controlled over batch sizes of new observations (incremental batch  learning), with a batch size of $1$ corresponding to online learning.

\section{Lambda Learning}\label{Lambda Learning}
% Section 3
% It is unclear why you are doing both steps 14 and 15 on full Hessian and the diagonal. I am searching for the discussion on the two methods, but I am unable to find it in section 3.
In this section, we present our incremental learning framework. Our approach provides an efficient mechanism for updating the random-effects model on fresh data flows. One key aspect is the parallelization of random-effect model training, which is proved possible when data instances are grouped by random-effect types \citep{zhang2016glmix}.

\subsection{Formulation for Incremental Updates}
For every batch of data $\Omega^{(t)}$, we want to solve the problem
\begin{alignat}{1}
\label{glmix:main1}
\minimize{\beta} &\,\,\,\, F_t(\beta) - \log p(\beta),
\end{alignat}
where $p(\beta)$ is the prior distribution of the model parameters $\beta$. Assuming a Gaussian prior $p(\beta) \sim {\mathcal{N}} (0,\lambda I)$, as in the case of log-logistic loss, the output of the problem is a Gaussian posterior distribution $p_t(\beta) \sim {\mathcal{N}}(\beta^{*(t)},\Sigma^{*(t)})$ (using Laplace approximation).

However, as the data grows, it becomes unfeasible to store in memory, and too computationally expensive to completely iterate over to compute $F_t$, $\nabla F_{t}$, and/or $\nabla^{2} F_{t}$. Therefore, we maintain an approximation of the loss function over the past observations, $F_{t-1}$, and combine it with the loss function over the latest batch $f_t$. In particular, we decompose (\ref{glmix:main1}) into a loss on past data and a loss on an incremental batch,
\begin{alignat}{1}
\label{glmix:main2}
\minimize{\beta} &\,\,\,\, \delta F_{t-1}(\beta)  + f_t(\beta) - \log p(\beta),
\end{alignat}
We substitute the first term of the objective function though approximation using a quadratic function centered around previous optimum $\beta^{*(t-1)}$,
\begin{alignat}{1}
\label{glmix:main3}
\minimize{\beta} &\,\,\,\,   \tfrac{\delta}{2}\|\beta - \beta^{*(t-1)}\|^{2}_{\nabla^{2} F_{t-1}(\beta^{*(t-1)})} + f_t(\beta) + \tfrac{\lambda}{2}\|\beta\|^{2}.
\end{alignat}
Note that optimality of $\beta^{*(t-1)}$ implies $\nabla F_{t-1}(\beta^{*(t-1)}) = 0$ (hence the linear term is dropped).

In the Bayesian framework, since $\left(\Sigma^{*(t-1)} \right)^{-1} = \nabla^{2}F_{t-1}(\beta^{*(t-1)})$, this approximate problem is equivalent to training the mini-batch using the posterior distribution, $\mathcal{N}(\beta^{*(t-1)},\Sigma^{*(t-1)})$, from the previous step as a prior distribution for $\beta^{(t)}$.

\subsection{Asynchronous Updates}\label{asynchronous-formulation}
Additionally, in the GAME framework, we can use asynchronous updates for the random-effect models to keep them fresh in production systems. We control the rate at which a random-effect model is updated by a mini batch-size parameter. Once a sufficiently-sized batch of data $\Omega^{(t)}_{\bar{r}\bar{l}}=\{ n \in \Omega^{(t)} | i(\bar{r},n)=\bar{l} \}$, for random-effect $\bar{r}$ and index $\bar{l}$, has been observed, we update the corresponding random-effect by solving:
\begin{alignat}{1}
\label{glmix:incremental-batch}
\minimize{\beta_{\bar{r}\bar{l}}^{(t)}} \ & \sum_{n \in \Omega^{(t)}_{\bar{r}\bar{l}}} - \log \left(y_n | \zeta^{(t-1)}_n + {\bf z}_{\bar{r}n}^{T}\beta_{\bar{r}\bar{l}}^{(t)} \right) - \delta \log p(\beta_{\bar{r}\bar{l}}^{(t)}) \nonumber \\
& \zeta^{(t-1)}_n = {\bf x}_n^T\beta^{*(0)}_f + \sum_{s \in {\mathcal{ R}}, r \neq \bar{r}} {\bf z}_{rn}^{T}\beta_{r,i(r,n)}^{(t-1)},
\end{alignat}
and the posterior distribution for $\beta_{\bar{r}\bar{l}}^{(t-1)}$, i.e., the Gaussian distribution $\mathcal{N}(\beta_{\bar{r}\bar{l}}^{*(t-1)}, \Sigma^{*(t-1)}_{\bar{r}\bar{l}}$) is used as the prior distribution $p(\beta_{\bar{r}\bar{l}}^{(t)})$. We keep the fixed-effect model, $\beta^{*(0)}_{f}$, stationary throughout the incremental training process and use the updated model $\beta^{*(t)} = [\beta^{*(0)}_{f}, \{B^{*(t)}_{r} \}_{ r \in {\mathcal{R}}}]$ to serve traffic after time $t$. The fixed-effect model ($\beta^{*(0)}_{f}$), typically an ensemble of embedding based models and boosted forests, are continuously retrained offline.

\subsection{Algorithm Details}
Algorithm 1 describes the complete algorithm. We start with large volumes of historical engagement data to train, in offline fashion, a fixed-effects (ensemble) model as the initial batch of random-effects models. This produces fixed-effects coefficients (\textit{generalization} part of the model), $\beta^{*(0)}_{f}$, as well as a starting point $\beta^{*(0)}_{r} = \beta^{*(0)}_{r,inc}$ for incrementally learning the random-effect (\textit{memorization} part of the model) coefficients. This is also used to set the initial Hessian matrix, $H^{(0)}_{r} = \nabla^{2}{ F_0}({\beta_r^{*(0)}})$ as well as the time discount $\delta < 1$. Every time the fixed-effects models are retrained offline, all of the above parameters are asynchronously updated from outside the system.

Now, as the system processes the next mini-batch of data, it uses the latest coefficients from the fixed-effect model to compute an offset for each training example. Note the same stationary $\beta^{(0)}_{f}$ for all $t$.  The offset provides the connection of the fixed-effect models with the random-effects that are to be updated while allowing different model families to be used for generalization and memorization. We then train the random-effect coefficients ($\beta^{*(t)}_{r,inc}$) using the current batch of data while constraining the solution to be ``near'' the previous optimum (line 13 of Algorithm 1). In practice, the entire training loop for each random-effect executes in under a second. The new coefficients are used to update the running Hessian estimate and the posterior covariance. The model coefficients, $\beta^{(t)} = [\beta_f^{*(0)}, \beta_r^{(t)}]$, are then pushed to stores for online serving. Since we have access to the posterior covariance, we rely on $\beta^{(0)}_{f}$ and perform Thompson Sampling from the updated distribution for $\beta^{(t)}_{r}$ to compute scores online.

\textbf{Connection to Reinforcement Learning:} The ad ranking problem is often cast as an instance of a contextual multi-armed bandit problem\cite{langford2007epoch}. Each of the possible ads is represented by an arm ($r \in \mathcal{R})$. The action space (arms) is dynamic as ads get introduced into the system when they are created and retired as the budget runs out. The context for opportunity $t$ is encoded as a feature vector ${\bf x_f}$. The policy uses the context to make a decision at every opportunity (a member request) to serve an ad. Once a decision is made an observation is collected, and the corresponding reward computed; e.g., member clicked on the ad. The policy is then revised with the data collected for this opportunity. An effective way to solve this bandit problem is via Thompson sampling\cite{chapelle2011empirical,tang2013automatic}. Since our algorithm produces the posterior covariance as an output, the system behaves like an online contextual bandit. As new arms get introduced, the system is able to estimate to value of each arm much faster because of the fast feedback loop. Additionally, reducing the number of high variance opportunities for new ads improves the auction efficiency and overall revenue.

\section{Theory}\label{Theory}
% Section 4
% The theory is weak and also not brought out well. 
In this section, we give theoretical guarantees of why incremental training improves model performance. To start, we prove that the first incremental step at $t=1$ is better and then extend it to subsequent steps $t=2, 3, \ldots$.
\subsection{First incremental update ($t=1$)}
Starting with a exact solution $\beta^{*(0)}$ trained on the initial batch $\Omega^{(0)}$, we make an incremental update based on the first mini-batch of observations $\Omega^{(1)}$.
\begin{theorem}
\label{theo1}
For a convex loss function $f$, the first incremental step update $\beta^{*(1)}_{inc}$ obtained by solving problem (\ref{glmix:main3}) for $t=1$, i.e.,
\begin{alignat}{1}
\label{th1}
\beta^{*(1)}_{inc} &= \argmin{\beta} \,\,\,\, f_1(\beta) + \tfrac{\delta}{2}\|\beta - \beta^{*(0)}\|^{2}_{\nabla^{2} F_{0}(\beta^{*(0)})},
\end{alignat}
improves the loss on $\bar{\Omega}^{(1)}$, i.e. $F_1(\beta)$, compared to the previous batch update $\beta^{*(0)}$. In other words,
\begin{alignat}{1}
\label{th2}
F_{1}(\beta^{*(0)}) - F_{1} (\beta^{*(1)}_{inc}) \geq \tfrac{\lambda}{2} \|\beta^{*(0)}-\beta^{*(1)}_{inc})\|^2_2 + {\mathcal O}(\gamma^3),
\end{alignat}
where, $\gamma = \|\beta^{*(1)}_{inc}-\beta^{*(0)})\|_{2}$.
\end{theorem}

\subsection{Subsequent incremental updates ($t=2,\ldots$)}
We begin by defining a series of incremental objective functions (and respective optimas as incremental updates) similar to (\ref{gen:main}),
\begin{alignat}{1}
\label{inc:main}
F_{t}^{inc}(\beta) &= f_t(\beta) + \tfrac{1}{2}\|\beta - \beta_{inc}^{*(t-1)}\|^{2}_{\nabla^{2} F^{inc}_{t-1}(\beta_{inc}^{*(t-1)})}, 
\end{alignat}
and show that these incremental updates trained on mini-batches are better using the stale model parameters for any $t$.
\begin{theorem}
\label{theo2}
For a convex loss function $f$, for $t = 2, 3, \ldots$, the series of incremental updates $\beta^{*(t)}_{inc}$ obtained by solving problem (\ref{inc:main}) for $t=2, 3, \ldots$ , improves the loss compared to the stale batch update $\beta^{*(0)}$. In other words,
\begin{alignat}{1}
\label{th21}
F_{t}(\beta^{*(0)}) - F_{t}(\beta_{inc}^{*(t)}) \geq \tfrac{\lambda}{2} \|\beta^{*(0)}-\beta^{*(t)}_{inc} \|^{2}_2 + {\mathcal O}(\bar{\gamma}^3),
\end{alignat}
where $\bar{\gamma} = \max \left\{ \|\beta^{*(k)}_{inc} - \beta^{*(t)}_{inc}\|_2, \|\beta^{*(k)}_{inc} - \beta^{*(0)}\|_2 \right\}$, for $0 \leq k \leq t$.
\end{theorem}

\noindent \textbf{Note}: The optimality gap between the loss-function values depend on the distances between $\beta^{*(0)}$ and $\beta^{*(t)}_{inc}$, which in turn depend on the drift in distributions of the data set between $\Omega^{(0)}$ and subsequent $\Omega^{(t)}$'s.

\section{System Overview}\label{System Overview}
This section describes a baseline system capable of offline retraining and online scoring, and how we augment it to support high-frequency nearline updates. Any system using the GAME paradigm can adopt our solution and benefit from improved model freshness, and resulting lifts to publisher ROI and content engagement.

\subsection{Motivation}

    We would ideally like to update a time-sensitive model as soon as new observations are available, with zero delay. A typical system, where offline batch updates occur with period $T$, is insufficient. The data size itself increases with $T$, since a new batch of data accumulates while the model is trained on the previous batch. In practice, we observe $T$ can be 8-24 hours due to the time needed to prepare a new training data batch and retrain on this batch.

    Such offline batch retraining is common practice for GAME (fixed + random effects) models (\ref{glmix:main3}). The promise of \textit{Lambda Learning} is to reduce the time to adapt to new observations by processing data incrementally and learning on sequences of mini-batches nearline. This is a significant step towards the ideal of immediate model updates.

\subsection{Overview of sponsored content ad system}

\begin{figure*}
      \includegraphics[width=0.8\textwidth]{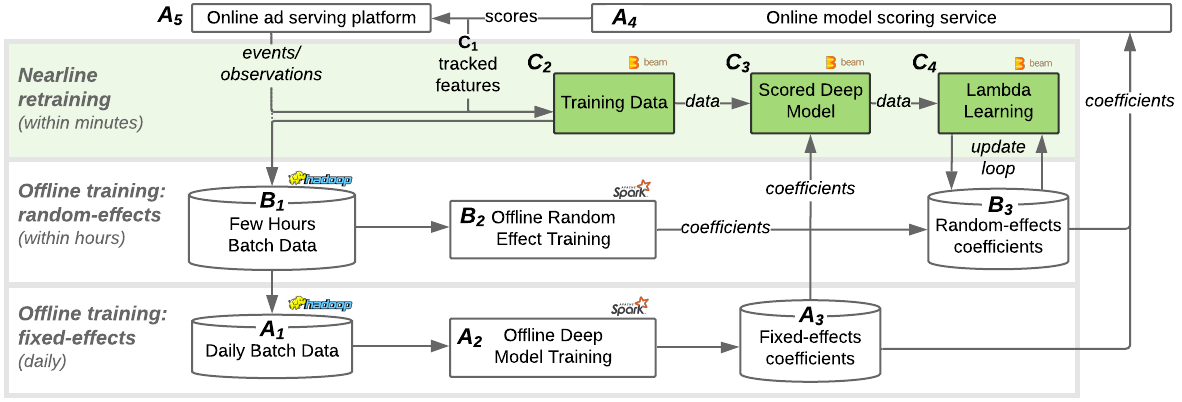}
      \vspace{-0.1in}
      \caption{System design for incremental nearline model updates. Nearline components are highlighted in green.}
      \label{fig:ads-architecture}
      \vspace{-0.15in}
    \end{figure*}

    Figure \ref{fig:ads-architecture} illustrates a baseline system capable of offline retraining and online scoring, and an extension, shown in green, which enables high-frequency nearline model updates. The components we introduce are: feature tracking, nearline training data generation, fixed-effect model scoring, and nearline random-effects model training (\textit{lambda learning}).
    
    Our solution operates in the context of a large sponsored content advertising system (A5 in Fig. \ref{fig:ads-architecture}). A machine learned response prediction model plays a central role, scoring content by estimating user engagement. The base production system can be logically separated into two phases: (1) a cold start phase where the fixed-effects part of the GAME model is trained (A in Fig. \ref{fig:ads-architecture}), and (2) a warm start phase where the random-effects coefficients are trained, boosting the scores of the fixed-effects model (B in Fig. \ref{fig:ads-architecture}). The fixed-effects part is typically a large non-linear model, trained on user attributes, activity, and contextual cues, and retrained periodically. The random-effects capture information specific to the ad, campaign, and advertiser. Random-effect models are small and linear, allowing modeling control at different levels of granularity. Random-effects of a given effect-type can be trained independently, allowing us to train only those for which there is new data (e.g. new ad content). They are also time-sensitive (especially for previously unseen advertisers/creatives), and are most effective when retrained frequently. See \cite{agarwal2014laser} for more background on this paradigm.
    
\subsection{Prerequisite system components}
    
    The baseline system is capable of offline retraining and online serving:

    \subsubsection{Offline Batch Model Training for $\Omega^{(0)}$}
    This phase involves training an initial GAME model, $\beta^{*(0)} = [\beta^{*(0)}_{f}, \{ \beta^{*(0)}_{r} \}_{ r \in {\mathcal{R}} }]$. In our case, we train a GAME model $\Omega^{(0)}$ over a training window of $D$ days (typically 14-30 days, depending on naturally occurring periodicity). Since $\beta^{(0)}_{f}$ represents a generalized fixed-effects model learned across a complete dataset, it is tolerant to minor data shifts and can be updated less frequently (in our case, once every $d$ days).  However, since new items are constantly added to the system and item-specific behaviors are fast-changing, $\{B_{r} \}_{ r \in {\mathcal{R}} }$ must be updated frequently, every $h$ hours (e.g. 6-8 hours) to capture these changes.

    \begin{tcolorbox}[size=small, after skip=6pt]
        \textbf{\emph{Step 1}}: For each training cadence of $d$ days, we start fresh and train $\beta^{*(0)} = [\beta^{*(0)}_{f}, \{ \beta^{*(0)}_{r} \}_{ r \in {\mathcal{R}} }]$ on the observations, $\Omega^{(0)}$, collected over training window $T$ by solving Equation (\ref{glmix:main}).
    \end{tcolorbox}
    
    % \textbf{\emph{Step 2}} Every $h$ hours, we update a subset of coefficients, the random effects $\{B_{r} \}_{ r \in {\mathcal R} }$, by solving \ref{glmix:random}. In practice, we solve this by cyclically iterating over each type of random effect $r$. For a given random effect $\bar{r}$, we assume all other coefficients to be held constant, the optimization problem further reduces to a separable set of smaller sub-problems given by
    % \begin{alignat}{1}
    % \label{glmix:sub-random}
    % \maximize{\beta_{\bar{r}}} \ & -\sum_{n \in \Omega_{\bar{r}}} \log \left(y_n | \zeta^{(0)}_n + {\bf z}_{\bar{r}n}^{T}\beta_{\bar{r},i(\bar{r},n)} \right) +  \sum_{l=1}^{N_{\bar{r}}} \log p(\beta_{\bar{r}l})
    % \end{alignat}
    % here, $\displaystyle\zeta^{(0)}_n = {\bf x}_n^T\beta^{(0)} + \sum_{r \in {\mathcal R}, r \neq \bar{r}} {\bf z}_{rn}^{T}\beta_{r,i(r,n)}^{(0)}$.

    \subsubsection{Online model serving (A4 in Fig. \ref{fig:ads-architecture})} The random-effects are trained on varying volumes of training data, resulting in different covariance estimates for the $\beta_r$'s. We use Thompson Sampling to incorporate these covariance estimates in online inference, enabling exploration of low-mean, high-variance ads \cite{agarwal2014laser}.
   
    \begin{tcolorbox}[size=small, after skip=6pt]
        \textbf{\emph{Step 2}}: For each of the random-effects coefficients, we use the posterior mean $\beta_{\bar{r}}^{*(0)}$, and the variance $\Sigma_{\bar{r}}^{*(0)}$ to sample a $\beta_{\bar{r}}$ for online inference and serving. 
    \end{tcolorbox}
    
\subsection{System components for Lambda Learning}
 
    To extend the baseline system to support frequent nearline model updates, we add four components: feature tracking, nearline training data generation, fixed-effect model scoring, and nearline random-effect model training (\textit{lambda learning}). These communicate via Kafka topics partitioned by random-effect id. Horizontal scaling is achieved by increasing the number of partitions and machines running these components.
 
    \subsubsection{Feature Tracking (C1 in Fig. \ref{fig:ads-architecture})}
    
    Common practice in model training pipelines is to join labeled data instances (``observations'') with offline stores of uniquely keyed features (e.g. member features, ad features, etc.) to produce training data, $\Omega$.  The trained models are then applied on-demand in an online serving layer, where ``feature producer'' functions produce a real-time representation of ${\bf x}_n$ prior to being scored with the model. However, we observe drawbacks to this approach: (1) differences between offline/online feature production (e.g. due to bugs, versioning, unavailability, etc), and (2) time-sensitivity of certain feature sets. Training pipelines that rely on the present value of time-sensitive features (e.g. activity-based features) can cause overfitting, and, in extreme instances. introduce label information into training data.
    
    As a solution, we introduce an online feature tracking system which records feature values used in online scoring. This is implemented using Kafka \cite{kreps2011kafka}, an distributed messaging system, which allows services to asynchronously share data in near real-time. This solves the issue of feature time-sensitivity, since feature values are computed at the same time as the observations from which we create training labels. For online services that already score and fetch features on-demand, tracking these features takes little extra work. For efficiency, we track only time-sensitive features, joining on static features in the next component.
    
    \subsubsection{Training Data Generation (C2 in Fig. \ref{fig:ads-architecture})}
    
    Kafka provides \textit{topics} such as impressions, clicks, and tracked features. We use Samza \cite{noghabi2017samza} to listen to these topics and produce featurized training examples, which are published to a new Kafka topic. This data is used by subsequent nearline training, but is also ETLed offline for use in offline training. Online/offline feature parity is guaranteed. Features that are fairly static, but dense (e.g. embeddings of static entities) are good candidates to be introduced at this stage since it not cost efficient to serialize highly redundant, dense features.
    
    Two important knobs that are fine-tuned by analysing traffic patterns for data quality are:
    \begin{itemize}
    \item \textbf{Trade-off: training cost and accuracy}: Since offline data pipelines have access to ``delayed'' clicks when preparing training data, all clicks are accounted for. In the streaming case, the use of windowing means ``late'' click may be missed. Longer windows mean higher cost, but lower rate of misses.
    \item \textbf{Partition for throughput}: Topics are partitioned based on random-effect id. This avoids network shuffle, and ensures all data for a given advertiser is processed on the same node.
    \end{itemize}
    
    \subsubsection{Fixed Effect Model Scoring (C3 in Fig. \ref{fig:ads-architecture})}
    
    This Samza processor listens to a stream of training data and scores each incoming instance with the fixed-effects model parameterized by $\beta_f^{(0)}$. If the fixed-effects model is an ensemble, the entire ensemble is scored just like it would be online to compute a score for every impression. The score serves as the offset for random-effects training as explained in Algorithm \ref{glmix:logloss}.
    
    \begin{tcolorbox}[size=small, after skip=6pt]
        \textbf{\emph{Step 3}}: Compute the score from the fixed-effects model, $g_f(x_n, \beta_f^{(0)})$.
    \end{tcolorbox}
    
    \subsubsection{Lambda Learning (C4 in Fig. \ref{fig:ads-architecture})}
    
    The random-effects coefficients produced by offline training are served online via a key-value store. The last step, Lambda Learning, involves iterative retraining of these using the nearline data. This component groups stream data into mini-batches based on random-effect id, fetches the corresponding random-effects coefficients from the store, updates them, and writes them back, ready for online scoring.
    
    In practice, the velocity of incoming data streams can fluctuate throughout a day. Therefore, mini-batch updates can be triggered by meeting one of many conditions, such impression count (high velocity periods), time since last update (low velocity periods), or coefficient variance (new ids).
    
    \begin{tcolorbox}[size=small, after skip=6pt]
        \textbf{\emph{Step 4}}: For each mini-batch of data $\Omega^{(t)}$, we solve the inner loop of Algorithm \ref{glmix:logloss} to update $\beta^{*(t)}_{r}$ and $\Sigma^{*(t)}_{r}$.
    \end{tcolorbox}
    
    Several subtleties arise in implementing the read-train-write (RTW) part of this component:
    \begin{itemize}
    \item \textbf{Concurrence}: While different random-effects can be trained in parallel, RTWs for a given random-effect must be sequential. We enforce this by making each RTW hold a lock unique to the random-effect being retrained.
    \item \textbf{Weak store consistency}: Derived data key-value stores may offer weak consistency guarantees to meet strict latency or throughput SLAs. In particular, the store we use doesn't guarantee read-your-write semantics. We've found that placing a TTL cache between the processor and the store solves this issue.
    \item \textbf{Replay after batch updates}: The store supports both batch writes and nearline streaming writes. When a batch write occurs it is based on hours-old data. Replaying the training data from the last few hours on top of the new batch coefficients ensures coefficients reflect the most recent data.
    \end{itemize}
    
    % Additionally, since the incremental loss formulation is an approximation, the error can compound and degrade the model. It is important to find the sweet spot where we benefit from more frequent model updates without incurring too much compounded error before the offline model refreshes.

% \subsection{Wider Generalization}
% The Lambda Learning framework has a wide applicability outside of just Sponsored Updates and is currently being integrated across multiple product verticals that use the GAME family of models. Some design decisions that helped improve adoption are detailed below:
% \begin{itemize}
%     \item Modularity - This pattern of model training can be applied to any recommender system / ranking model, i.e. the existing model becomes the global part of the GAME model and we can always train a simple random effect model on the residual of the existing model. This allows for a higher degree of personalization and control.
%     \item Robustness - In case of model issues, i.e. convergence, overflow etc. it should be easy to rollback to previous snapshots of the model. Since the existing architecture overwrites the  coefficients to the online store, we design test model performance post training on the incoming stream as a sanity check before we trigger the write.
%     \item Backwards compatibility - It is designed to be retrofitted to existing patterns for online scoring and ranking. Applications that adopt this framework will not need change their architectures as long as they can track the features that are used to score the model.
% \end{itemize}

\section{Experiments}\label{Experiments}

%While analysing our framework, we would like to evaluate three scenarios: 1) a full batch update, 2) the incremental algorithm's update and 3) %the lambda update, i.e. full batch updates interspersed with incremental updates.

%As the new data becomes available for training we initiate model updates. Incremental algorithm has access to only the most recent data and %batch algorithm has access to all the data. Considering the system delays (e.g., nearline vs online), we consider two scenarios: instantaneous %update (ideal) and delayed update (realistic). In the ideal scenario model becomes available for scoring immediately. In the realistic scenario %there is a system overhead associated with data collection, model training and deployment.

We evaluate our work across two datasets, focusing on the performance comparison between the proposed incremental-online updates, and a traditional iterative, batched-offline approach. Our experiments, though conducted offline using static datasets, consider the online data stream from which they were sourced. The purpose of the experiments is to measure the long-term stability of incremental updates, as well as the immediate impact of including near real-time data in our models. Our hypothesis is that the incremental approach's increased access to recent data will increase its predictive power. This data advantage is proportional to the time-costs associated with iterative-offline training, and therefore, we believe incremental updates will be most effective as offline training time increases. However, we also hypothesize that prolonged incremental updating without iterative-offline training can compound the error associated with the incremental loss-function.

\subsection{Datasets}
We use two datasets: a public ``\textit{movie lens 20M}'' dataset \citep{harper2015movielens} for reproducibility and a proprietary sponsored updates dataset which is derived from the system we have built and used to train the ads response prediction models. All of our findings are consistent between the two datasets, so we present the detailed discussion of the results on the ads dataset. The corresponding results on the movie-lens dataset are in Appendix~\ref{appendix:movie}.

\subsubsection{Social network ad click data}
We collected a small number of weeks of ad click and impression data from a large social network. This time frame is sufficient to simulate life-cycle of an offline model and incremental lambda-learner model. We are omitting details about the features and transformations used for confidentiality reasons. However, the dataset includes sufficient information about the advertisements themselves and the audience of the impressions.

\subsubsection{Movie Lens}
Movie lens 20M \citep{harper2015movielens} is a benchmark dataset with 20 million ratings given to 27,000 movies by 138,000 users. We use this dataset for reproducibility of results and apply transformations to make the setup more analogous to that of our real-world ``\textit{sponsored updates}``) dataset: 1) We binarize responses, using scores greater or equal to 4.0 as positive labels and scores less than 4.0 as negative labels, 2) We use ALS matrix factorization \citep{matrixFactALS2009} of rank 30 to produce latent movie features used in the model, 3) As the dataset spans 20 years and is not as time sensitive as a typical web recommendation, we speed up time and scale 20 years into 14 days.

\begin{figure}
  \vspace{-0.1in}
  \centering
  \includegraphics[width=1.1\linewidth]{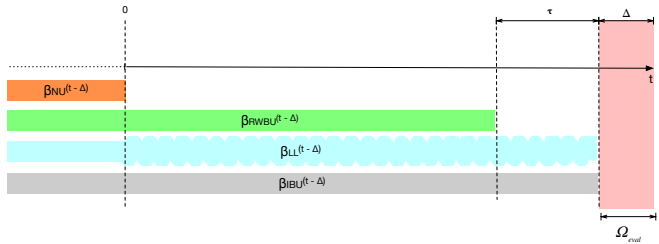}
  \vspace{-0.25in}
  \caption{Illustrates how models are evaluated. Timeline $[0,t)$ represents a time-ordered dataset. $\Delta$ represents the size of an evaluation set after each model update, $\Omega_{eval}$, and $\tau$ is an estimated delay incurred by full-batch training. $\beta_{NU}$, $\beta_{RWBU}$ and $\beta_{IBU}$ represent the parameterized models produced using full-batch training, and $\beta_{LL}$ represents parameterized models produced using incremental online updates (shown in waves). Incremental updates are executed every $\Delta$, and total number of incremental updates is given by $T/\Delta-1$}
  \label{evaluation-framework}
  \Description{Evaluation Framework}
  \vspace{-0.19in}
\end{figure}

\subsection{Evaluation Framework}
We chronologically order each of the datasets and split them into two parts. The first part of the data, $\Omega_{cold}$, is used to train an initial, baseline model, $\beta^{(0)}$. The latter part, $\Omega_{warm}$, is used to simulate model updates and for evaluation. Let us assume that  $\Omega_{warm}$ starts at time $t = 0$ and ends at time $t = T$. We break $\Omega_{warm}$ into $T/\Delta$ increments of equal duration, $\Delta$. The models, described below, are evaluated sequentially on all the increments. After an increment is used for evaluation, it becomes available for training (with a delay $\tau$, representing the cost of batch training and deployment).

Our work is motivated by the computational costs associated with offline iterative training of random-effect models. We compare the iterative approach against batch updates with different training costs, $\tau$. When $\tau = 0$, offline model has no associated cost, and updates happen instantaneously.  As $\tau \rightarrow \infty$, offline training becomes impossible, and model updates are infeasible. Considering this, we compare four model update variants (illustrated in Figure \ref{evaluation-framework}):

\begin{figure}
  \vspace{-0.1in}
  \begin{subfigure}{0.37\textwidth}
    \includegraphics[width=\textwidth]{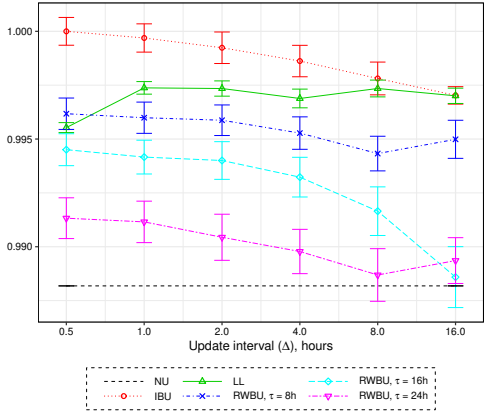}
    \caption{Change in ROC-AUC, as model update frequency is reduced. The x-axis denote the models being updated every $\Delta$ hours - in the incremental using ONLY data in the last $\Delta$ hours and in the batch using the entire history including the data in the last $\Delta$ hours.}
    \label{ads-batch}
  \end{subfigure}
  \hfill
  \begin{subfigure}{0.37\textwidth}
    \includegraphics[width=\textwidth]{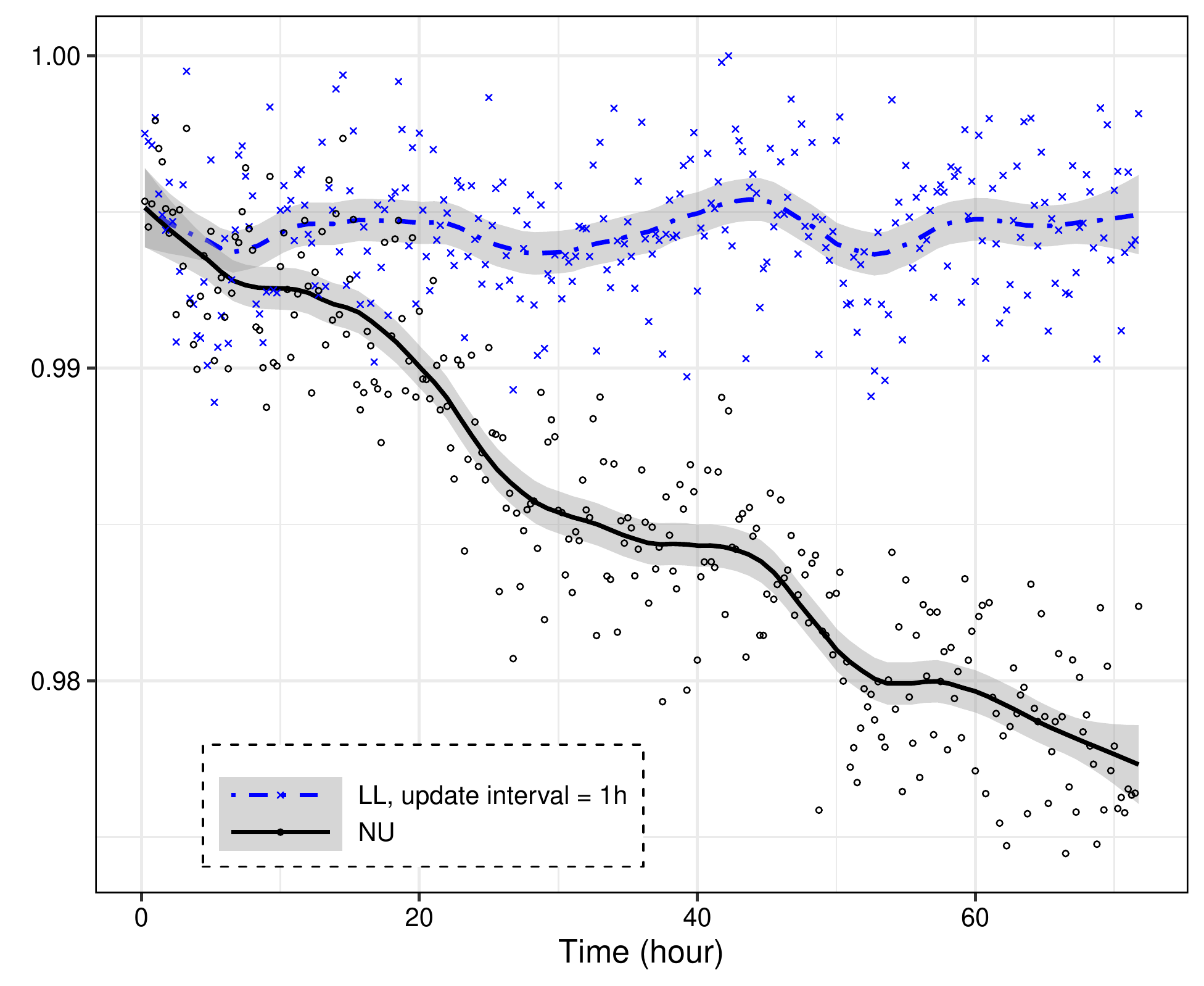}
    \caption{The best LL variant performance is stable and the batch performance degrades when the area under ROC metrics are computed every 15 minutes for a period of 72 hours. 95\% confidence intervals included.}
    \label{fig:ads_decay}
  \end{subfigure}
  \vspace{-0.1in}
  \caption{Model evaluation using lambda learning on ad click production data.  {\bf Note:} y-axis on both charts is scaled to obfuscate absolute metric values, highest value corresponds to 1.0. Historical weight was fixed, $\delta = 0.95$.}
  \vspace{-0.28in}
\end{figure}

\begin{itemize}
\item \textbf{No Updates (NU)}: Under this strategy, we assume $\tau = \infty$, and no model updates are performed. This means that all increments (starting at 0, $\Delta$, 2$\Delta$ ...), are evaluated using the model, $\beta^{(0)}$. This serves as a strict lower-baseline for other variants.
\item \textbf{Ideal Batch Update (IBU)}: To approximate the error of incremental model updates against an upper bound, we compare against an offline batch model trained without compute cost ($\tau$ = 0).
\item \textbf{Real-World Batch Update ($RWBU_{\tau}$)}: We evaluate full-batch updates over different training costs $\tau$. Each increment, $i\Delta$, is evaluated using, $\beta^{(i\Delta-\tau)}$, which is trained offline on all data instances preceding the $\Omega_{eval}$ by more than $\tau$.
\item \textbf{Lambda Learner ($LL_{\Delta}$)}: We compare against model updates based on the proposed method in Section \ref{Lambda Learning}. For LL, data availability is made identical to that of IBU. However, models are updated every $\Delta$. In practice, $\Delta$ should be small compared to the batch training cost $\tau$, as comparable or larger values imply updates as infrequent as RWBU.
\end{itemize}

Updated models are evaluated on each $\Delta$, using the area-under-ROC (AUC) \citep{hanley1982meaning} as the metric. We compute AUC on the full $\Omega_{warm}$ dataset (i.e. all $T/\Delta$ increments).

\subsection{Incremental vs. Batch Loss}
First, we quantify the drop in performance caused by the approximation of the incremental loss function. We measure this drop at different incremental update frequencies (i.e. different values of $\Delta$, shown in Figure \ref{ads-batch}). We report the aggregate metric for all model updates over $\Omega_{warm}$ to remove the effects of seasonality, intra/inter-day trends, and any other idiosyncrasies that may cause inaccuracies over a single window.

Figure \ref{ads-batch} compares the performance of LL against the theoretical upper bound (IBU, $\tau = 0$), lower bound (NU) and some real world scenarios (RWBU, $\tau \in \{8, 16, 24\}$). For sake of completeness we show the performance values of all variants for each setting of $\Delta$, however, in an online deployment, it is hard to observe RWBU performance for values of $\Delta \le 4$ hours. This is because the time taken to finish a full retrain of the batch algorithm (e.g. $\tau = 8$ hours) is likely to exceed this update interval. The performance of the IBU degrades as $\Delta$ increases providing evidence for the need to have more frequent model updates. This behavior is typical in systems with dynamic content or when there are temporal patterns of data access. LL demonstrates consistent performance for all update frequencies except $\Delta = 0.5$ hours. We believe that two opposite trends balance each other and result in this stable performance: firstly, less frequent updates increase the delay, a detrimental factor; secondly, with less frequent updates there is less dependence on objective function approximation, a beneficial factor.

The model training and deployment delay ($\tau$) has strong impact on the performance of the batch update algorithm. LL starts to outperform the batch update with the delay of 8 hours (RWBU, $\tau=8$h) and significantly outperforms the batch update with higher values of delay ($\tau = 16$h and $\tau = 24$h).

The results and observations discussed in this section for the proprietary sponsored updates dataset are consistent with the results on the public Movie Lens data. We report these results in the Appendix \ref{appendix:movie} for reproducibility.

\subsection{Time Decay Comparison}
\label{subsec:time-delay}
In this section we investigate how the performance of competing algorithms decays with time. We expect the batch algorithm to decay between consecutive updates, while LL may accumulate error with every update. The magnitude of this decay determines whether ``frequent enough'' RWBU updates are sufficient for optimal performance, or if incremental updating can improve the performance between RWBU updates. In particular, we look at the performance of the non-updating batch offline model (NU) and LL model updated every hour ($\Delta = 1$ hour) within 3 days. To remove seasonality factors such as time-of-day and day-of-week we averaged over multiple runs with random start times.

The NU model is trained using data up to time $t=0$ in Figure \ref{fig:ads_decay} and the model performance clearly degrades with time. LL, on the other hand, shows no significant degradation.

\subsection{Effect of Forgetting Factor ($\delta$)}
\begin{figure}[h]
  \begin{subfigure}{0.23\textwidth}
      \includegraphics[width=\linewidth]{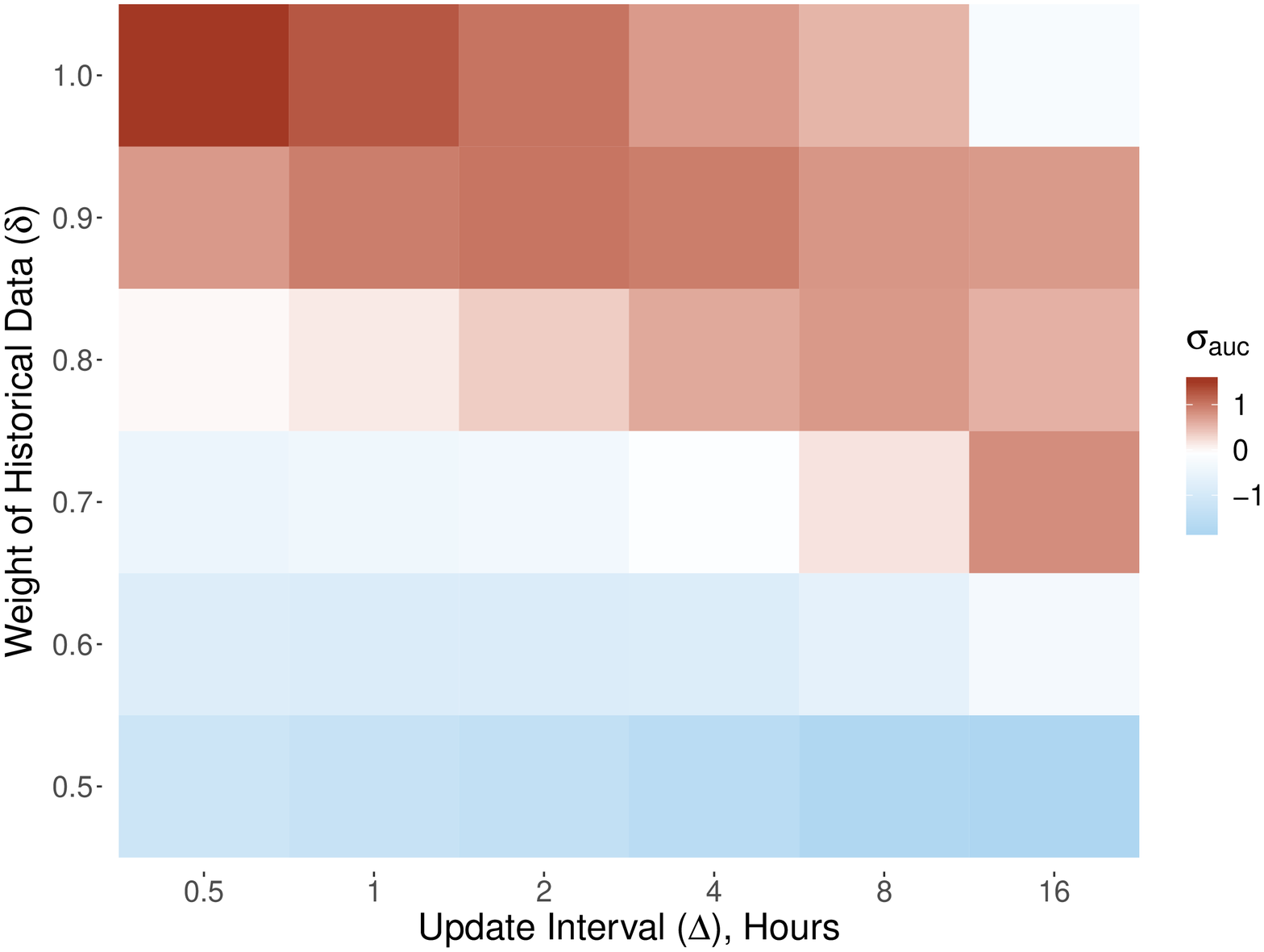}
      \caption{sponsored updates}
      \label{fig:delta-su}
  \end{subfigure}
  \hfill
  \begin{subfigure}{0.23\textwidth}
      \includegraphics[width=\linewidth]{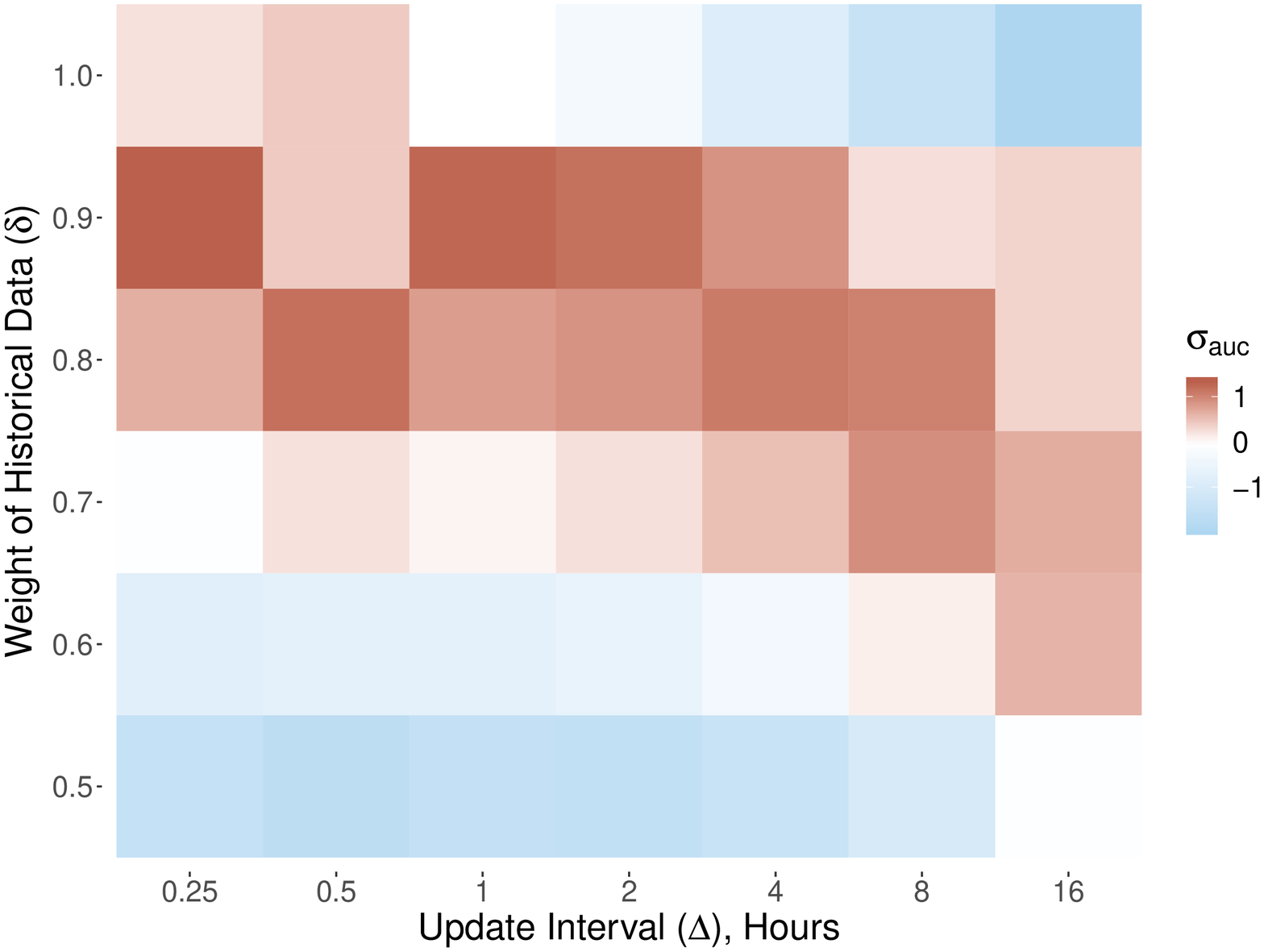}
      \caption{movie lens}
      \label{fig:delta-movielens}
  \end{subfigure}
  \vspace{-2mm}
  \caption{Heatmaps showing model performance against different $\delta$-values and update intervals ($\Delta$). Gradients are scaled against AUC values within a $\Delta$ value.}
  \label{fig:delta}
  \vspace{-3mm}
\end{figure}

The higher the frequency of an incremental model update, the smaller the batch of data used to update the random-effect models. With this, we find that the optimal $\delta$-values were inversely proportional to the batch size. Recall, as $\delta \rightarrow 0$ the updates focus more on the incremental data. Therefore, the smaller the batch size, the more we want to rely on the prior information learned and less on the incoming batch of data. On the contrary, if we have a large batch of data, then we have enough information to influence the Hessian in a meaningful way. We observed this trend on both datasets. Figure \ref{fig:delta} shows the trend of best performing $\delta$ for different update intervals, $\Delta$. All AUC values were scaled with respect to $(\mu_{auc}, \sigma_{auc})$, given a $\Delta$ value. It can be seen that when models are updated in shorter increments (smaller batches), then higher $\delta$ values shows better performance. However, as intervals grow longer (bigger batches), incorporating the new data becomes more effective.

\subsection{Effect of Hessian Approximation}
 On both datasets, we observe that model performance was not sensitive to using the full Hessian or its diagonal approximation to compute the prior. This could occur if the Hessian is well-conditioned. In production, we use the diagonal approximation for efficiency.

\section{Conclusion}\label{Conclusion}
Using the proposed nearline architecture, we close the feedback loop in the order of minutes. This reduces the number of high variance opportunities for new ads and allows Thompson sampling to converge faster. As a result new advertisers saw improvements in member engagement ($1.76\%$), ROI (reduced cost-per-click by $0.55\%$) and the platform saw an improvement in revenue ($2.59\%$). Existing advertisers also saw modest improvements across these key metrics resulting in an overall site-wide improvement in member engagement and platform revenue without hurting advertiser ROI.

It is a well-established industry practice to use the most recent user activity data to train machine learning models. Intuitively, in not doing so, one runs the risk of omitting time-sensitive information (e.g. trends, topics) that can better inform a recommendation system. That said, there are systemic delays (e.g. extract, transform, load data) and the training cost of machine learning algorithm (e.g. wide \& deep models) that limit the recency of data that can be used to update the model. To this end, our proposed \textit{Lambda Learning} framework trains the model on an incremental loss function using data prepared in near realtime. While the overall model is being periodically refreshed (batch retrain), \textit{Lambda Learner} continuously learns on an incremental objective and updates the model. This setup, along with the nearline system, improves the performance of the model serving production traffic. As shown in our experiments, the forgetting factor $\delta$ can be tuned to perform well on incremental batches of varying sizes. This also means that if we were to use custom $\delta_i$ for each random-effect (tuned based on the volume of data we've seen in the past), we can improve the model further.


% \begin{acks}
% We would like to thank Bee-Chung Chen, Sahin Geyik, Sathiya Keerthi, Hai Lu, David Pardoe, Shivani Rao and Ankan Saha for comments, feedback and thought-provoking discussions.
% \end{acks}

\bibliographystyle{ACM-Reference-Format}
\balance
\bibliography{lambda}

\newpage
\appendix
% \begin{align}
%     P(y) &= \frac{1}{1 + e^{-y x^T \theta}}\\
%     z &= \log (1 + e^{ x^T \theta}) \\
%     \log \mathbf{L}&= \frac{1}{2} (1 + y) \log P(y=1) + (1 - y) \log P(y=-1) \\
%     &= \frac{1}{2} [(1 + y) x^T\theta - 2z] = \frac{1}{2}(1 + y) x^T\theta -\log (1 + e^{ x^T \theta}) \\
%     \nabla{ \log \mathbf{L} } &= \left\{\frac{1}{2}(1 + y)  - \frac{ e^{x^T\theta} }{1 + e^{x^T \theta}}\right\} x  \\
%     \nabla{}^{2}{\log \mathbf{L}} &=  \left\{\frac{ -e^{x^T\theta} }{(1 + e^{x^T \theta})^{2}}\right\} x x^T\\
% \end{align}

\section{Algorithm}
\begin{algorithm}[H]
\caption{Lambda Learner - Sequential Online GAME Training}
\label{glmix:logloss}
\begin{algorithmic}[1]
\STATE \textbf{initialize}: 
\STATE \hspace{0.1in} \textit{Set prior distribution for} $\beta^{(0)} \sim {\mathcal{N}}\left(0, \lambda I\right)$,
\STATE \hspace{0.1in} \textit{Train global effects model on the initial batch $\Omega^{(0)}$}: \\
    \hspace{0.30in}  $\beta^{(0)}=[\beta^{*(0)}_{f}, \beta^{*(0)}_{r}]$, $\beta^{*(0)}_{r} = \beta^{*(0)}_{r,inc}$,\\
\STATE \hspace{0.1in} \textit{Set fixed-effects coefficient to $\beta^{*(0)}_{f}$}, 
\STATE \hspace{0.1in} \textit{Set posterior distribution for random-effects coefficients to} \\
 \hspace{0.30in} $\beta^{(0)}_{r} \sim {\mathcal{ N}}\left(\beta^{*(0)}_{r,inc},\Sigma^{*(0)}_{r,inc}\right)$,
\vspace{0.04in}
\STATE \hspace{0.1in} \textit{Set $H^{(0)}_{r} = \nabla^{2}{ F_0}({\beta_r^{*(0)}})$. Note, $\Sigma^{*(0)}_{r,inc} = \left(H^{(0)}_{r} + \lambda I\right)^{-1}$},
\vspace{0.04in}
\STATE \hspace{0.1in} \textit{Set time discount $\delta < 1$.}
\vspace{0.08in}
\FOR{time windows $t=1,2, \ldots$}  
    \STATE \textit{Prepare new batch of data:} $\Omega^{(t)}$,
    \STATE \textit{Compute fixed-effect offset:} $\zeta_{f} = {\bf x}_{f}^T\beta^{*(0)}_{f}, \forall {\bf x} \in \Omega^{(t)}$,
    \STATE \textit{Overall regularized log-logistic regression problem:}\\
    \ \ \ \ $ \minimize{\beta_r} \displaystyle \underbrace{ \delta F_{t-1}(\beta_r) }_{\substack{\textbf{time discounted}\\ \textbf{ log-loss on past}\\ \textbf{observations}}} + \underbrace{ f_t(\beta_r) }_{\substack{\textbf{log-loss on}\\ \textbf{most recent}\\ \textbf{observations}}} + \underbrace{\frac{\lambda}{2} \|\beta_r\|^{2}_2}_\textbf{regularization}$,
\vspace{0.1in}
    \STATE \textit{Approximate past log-logistic loss using quadratic approximation around $\mu_r^{(t-1)}$:} \\
    \ \ \ \  $ \displaystyle \delta F_{t-1}(\beta_r) =  \sum_{k=0}^{t-1} \delta^{t-k}  f_k(\beta_r) \approx \frac{\delta}{2} \|\beta_r - \beta_{r,inc}^{*(t-1)}\|^{2}_ {H_{r}^{(t-1)}}$,
\vspace{0.1in}
    \STATE \textit{Solve incremental logistic reg. to compute the posterior mean:}\\
    $\beta^{*(t)}_{r,inc} = \displaystyle \argmin{\beta_r} \ \frac{\delta}{2} \|\beta_r - \beta_{r,inc}^{*(t-1)}\|^{2}_ {H_{r}^{(t-1)}} + f_t(\beta_r) + \frac{\lambda}{2}\|\beta_r\|^{2}_2$,
\vspace{0.1in}
    \STATE \textit{Update for Hessian of approximated Log-logistic loss:} \\
    \ \ \ \  {$H_{r}^{{(t)}} = \delta H_r^{(t-1)} +  \displaystyle \sum_{{\bf x} \in \beta^{(t)}}$ $\frac{\exp{(\zeta_{f} + {\bf x}_{r}^{T}\beta^{*(t)}_{r,inc})}}{(1+\exp{(\zeta_{f} + {\bf x}_{r}^{T}\beta^{*(t)}_{r,inc})})^{2}}{\bf x}_r{\bf x}_r^T$},
% \vspace{0.1in}
%     \STATE \textit{Update for diagonal Hessian approximation:} \\
%     \ \ \ \  {$h_{r,(i,i)}^{{(t)}} = \delta h_{r,(i,i)}^{(t-1)} +  \displaystyle \sum_{{\bf x} \in \beta^{(t)}}$ $\frac{\exp{(\zeta_{f} + {\bf x}_{r}^{T}\beta^{*(t)}_{r,inc})} \ {x}_{r,(i,i)}^2}{(1+\exp{(\zeta_{f} + {\bf x}_{r}^{T}\beta^{*(t)}_{r,inc})})^{2}}$},
\vspace{0.05in}
    \STATE \textit{Compute the posterior covariance:} \\
    \ \ \ \ $\Sigma_{r,inc}^{*(t)} = \left( {H_{r}^{(t)} + \lambda I} \right)^{-1}$, %{(Using SVD)}}.
\vspace{0.05in}
    \STATE \textit{Update the model coefficients:} \\
    \ \ \ \  $\beta^{(t)} = [\beta_f^{*(0)}, \beta_r^{(t)}]$.
\vspace{0.05in}
    \STATE \textit{Serve online recommendations using Thompson sampling from updated distribution $\beta^{(t)}_{r} \sim {\mathcal{N}} \left(\beta^{*(t)}_{r,inc},{\Sigma^{*(t)}_{r,inc}} \right)$.}
\ENDFOR
\end{algorithmic}
\end{algorithm}

\section{Optimality Gaps for incremental updates}
In this section, we provide proofs of the two theorems. Beginning with a Lemma
\begin{lemma}
\label{lm1}
The Hessian for $F_{t}^{inc}(\beta)$ at $\beta=\beta^{*(t)}_{inc}$ is given by
\begin{alignat}{1}
\nabla^{2}F_{t}^{inc}(\beta^{*(t)}_{inc}) = \sum_{k=0}^{k=t} \delta^{t-k} \nabla^{2}f_{k}(\beta^{*(k)}_{inc}).
\end{alignat}
\end{lemma}
\begin{proof}
Follows by mathematical induction on Hessian of (\ref{inc:main}).
\end{proof}

\subsection{Proof of Theorem \ref{theo1}}
% \begin{theorem}
% For any convex loss function $f$, the first incremental step update $\beta^{*(1)}_{inc}$ obtained by solving problem (\ref{glmix:main3}) for $t=1$, i.e.,
% \begin{alignat}{1}
% \label{th1}
% \beta^{*(1)}_{inc} &= \argmin{\beta} \,\,\,\, f_1(\beta) + \frac{\delta}{2}\|\beta - \beta^{*(0)}\|^{2}_{\nabla^{2} F_{0}(\beta^{*(0)})}.
% \end{alignat}
% is better than using the previous batch update $\beta^{*(0)}$. In other words,
% \begin{alignat}{1}
% \label{th2}
% F_{1}(\beta^{*(0)}) - F_{1} (\beta^{*(1)}_{inc}) \geq \frac{\lambda}{2} \left\|\beta^{*(1)}_{inc}-\beta^{*(0)})\right\|^2 + {\cal O}(\gamma^3),
% \end{alignat}
% where $\gamma = \left\|\beta^{*(1)}_{inc}-\beta^{*(0)})\right\|_{2}$.
% \end{theorem}
\begin{proof}
We start with the recursive definition of $F_1$,
\begin{alignat}{1}
\label{pf1}
 F_{1}(\beta^{*(0)}) - F_{1} (\beta^{*(1)}_{inc}) &= \delta \left( F_{0}(\beta^{*(0)}) - F_{0} (\beta^{*(1)}_{inc}) \right) \nonumber\\
 & \ \ \ \ + f_{1}(\beta^{*(0)}) - f_{1} (\beta^{*(1)}_{inc}).
\end{alignat}
Using Taylor approximation around $F_0(\beta^{*(0)})$, since $\nabla F_0(\beta^{*(0)}) = 0$ due to optimality of $\beta^{*(0)}$, the first term reduces to
\begin{alignat}{1}
\label{pf2}
\delta \left( F_{0}(\beta^{*(0)}) - F_{0} (\beta^{*(1)}_{inc}) \right) = - \tfrac{\delta}{2} \|\beta^{*(1)}_{inc}-\beta^{*(0)})\|^2_{\nabla^{2}F_0(\beta^{*(0)})} + {\mathcal O}(\gamma^3).
\end{alignat}

\noindent Similarly, using Taylor's approximation around $f_1(\beta^{*(1)}_{inc})$, the second term reduces to
\begin{alignat}{1}
\label{pf3}
 f_{1}(\beta^{*(0)}) - f_{1} (\beta^{*(1)}_{inc}) & = \nabla f_1(\beta^{*(1)}_{inc})^{T} \left(\beta^{*(0)}) - \beta^{*(1)}_{inc}\right) \nonumber\\
 & \ \ \ \ +  \tfrac{1}{2} \|\beta^{*(1)}_{inc}-\beta^{*(0)}\|^2_{\nabla^{2}f_1(\beta^{*(1)}_{inc})} + {\mathcal O}(\gamma^3),
\end{alignat}
Next, using the optimality of $\beta^{*(1)}_{inc}$ gives 
\begin{alignat}{1}
\label{pf4}
\delta \nabla^2 F_0(\beta^{*(0)}) \left(\beta^{*(1)}_{inc} - \beta^{*(0)}\right) + \nabla f_1(\beta^{*(1)}_{inc}) = 0. 
\end{alignat}
Substituting equations (\ref{pf4}) in (\ref{pf3}), and combining with (\ref{pf2}) into (\ref{pf1}), we get  
\begin{alignat}{1}
\label{pf5}
F_{1}(\beta^{*(0)}) - F_{1} (\beta^{*(1)}_{inc}) &= \tfrac{1}{2} \|\beta^{*(1)}_{inc}-\beta^{*(0)})\|^2_{\nabla^{2} F_{1}^{inc} (\beta^{*(1)}_{inc})} + {\mathcal O}(\gamma^3),
\end{alignat}
where $F_1^{inc}(\beta)$ is the objective function in incremental problem (\ref{th1}), and $\nabla^{2}F_{1}^{inc}(\beta^{*(1)}_{inc}) = \delta \nabla^{2} F_0(\beta^{*(0)}) + \nabla^{2}f_1(\beta^{*(1)}_{inc})$ is the Hessian of that objective function. Therefore, for a convex loss function $f$, we have
\begin{alignat}{1}
\label{pf6}
F_{1}(\beta^{*(0)}) - F_{1} (\beta^{*(1)}_{inc}) & \geq \tfrac{\delta{m_0} + {m_1}}{2} \|\beta^{*(1)}_{inc}-\beta^{*(0)})\|^2 + {\mathcal O}(\gamma^3),
\end{alignat}
where $m_0$ and $m_1$ correspond to strong convexity constants of $F_0$ and $f_1$ respectively. And, in the Algorithm \ref{glmix:logloss}, $\delta{m_0} + {m_1} \geq \lambda > 0$, thereby ensuring strict improvement.
\end{proof}

\begin{lemma}
\label{lm1}
The Hessian for $F_{t}^{inc}(\beta)$ at $\beta=\beta^{*(t)}_{inc}$ is given by
\begin{alignat}{1}
\nabla^{2}F_{t}^{inc}(\beta^{*(t)}_{inc}) = \sum_{k=0}^{k=t} \delta^{t-k} \nabla^{2}f_{k}(\beta^{*(k)}_{inc}).
\end{alignat}
\end{lemma}
\begin{proof}
Follows by mathematical induction on Hessian of (\ref{inc:main}).
\end{proof}

\subsection{Proof of Theorem \ref{theo2}}
% \begin{theorem}
% For any convex loss function $f$, for $t = 2, 3, \ldots$, the series of incremental step updates $\beta^{*(t)}_{inc}$ obtained by solving problem (\ref{inc:main}) for $t=2, 3, \ldots$, is better than using the previous full update $\beta^{*(0)}$. In other words,
% \begin{alignat}{1}
% \label{th21}
% F_{t}(\beta^{*(0)}) - F_{t}(\beta_{inc}^{*(t)}) \geq \frac{\lambda}{2} \left\|\beta^{*(0)}-\beta^{*(t)}_{inc} \right\|^{2} + {\cal O}(\bar{\gamma}^3),
% \end{alignat}
% where $\bar{\gamma} = \max \left\{ \left\|\beta^{*(k)}_{inc} - \beta^{*(t)}_{inc}\right\|_2, \left\|\beta^{*(k)}_{inc} - \beta^{*(0)}\right\|_2 \right\}$, for $0 \leq k \leq t$.
% \end{theorem}

\begin{proof}
We start with base definition of $F_t$,
\begin{alignat}{1}
\label{pf21}
 F_{t}(\beta^{*(0)}) - F_{t} (\beta^{*(t)}_{inc}) &= \sum_{k=0}^{k=t} \delta^{t-k} \left( f_{k}(\beta^{*(0)}) - f_{k} (\beta^{*(t)}_{inc}) \right).
\end{alignat}
Using Taylor's approximation around $f_{k} (\beta^{*(k)}_{inc})$, we get
\begin{alignat}{1}
\label{pf22}
 f_{k}(\beta^{*(0)}) - f_{k} (\beta^{*(t)}_{inc}) &= \nabla f_k(\beta^{*(k)}_{inc})^{T} \left(\beta^{*(0)} - \beta^{*(t)}_{inc} \right) \nonumber \\ 
 & \ \ \ + \tfrac{1}{2} \|\beta^{*(0)}-\beta^{*(k)}_{inc}\|^2_{\nabla^{2} f_{k}^{inc} (\beta^{*(k)}_{inc})} \nonumber \\
 & \ \ \ - \tfrac{1}{2} \|\beta^{*(t)}_{inc}-\beta^{*(k)}_{inc}\|^2_{\nabla^{2} f_{k}^{inc} (\beta^{*(k)}_{inc})} + {\mathcal O}(\bar{\gamma}^3),
\end{alignat}
where $\bar{\gamma} = \max \left\{ \|\beta^{*(k)}_{inc} - \beta^{*(t)}_{inc}\|, \|\beta^{*(k)}_{inc} - \beta^{*(0)}\| \right\}$ for $0 \leq k \leq t$.
Using the optimality of $\beta^{*(k)}_{inc}$ gives 
\begin{alignat}{1}
\label{pf23}
\delta \nabla^2 F^{inc}_{k-1}(\beta_{inc}^{*(k-1)}) \left(\beta^{*(k)}_{inc} - \beta_{inc}^{*(k-1)}\right) + \nabla f_k(\beta^{*(k)}_{inc}) = 0.
\end{alignat}
Substituting (\ref{pf23}) in the first term of (\ref{pf22}), and using $\nabla f_0(\beta^{*(0)}_{inc}) = \nabla F_0(\beta^{*(0)}) = 0$ and Lemma \ref{lm1}, we get
\begin{alignat}{1}
\label{pf24}
&\sum_{k=0}^{k=t} \delta^{t-k} \nabla f_k(\beta^{*(k)}_{inc})^{T} \left(\beta^{*(0)} - \beta^{*(t)}_{inc} \right) \nonumber \\
&= - \sum_{k=1}^{k=t} \delta^{t-k} \left(\beta^{*(0)} - \beta^{*(t)}_{inc} \right) ^{T} \delta\nabla^2 F^{inc}_{k-1}(\beta_{inc}^{*(k-1)}) \left(\beta^{*(k)}_{inc} - \beta_{inc}^{*(k-1)}\right) \nonumber \\
&= - \sum_{k=1}^{k=t} \delta^{t-k} \left(\beta^{*(0)} - \beta^{*(t)}_{inc} \right) ^{T} \left(\sum_{j=0}^{j=k-1} \delta^{k-j} \nabla^{2}f_{j}(\beta^{*(j)}_{inc})\right) \left(\beta^{*(k)}_{inc} - \beta_{inc}^{*(k-1)}\right)
\end{alignat}
Now, switching the summations over $k$ and $j$, and using telescopic sum, we get
\begin{alignat}{1}
\label{pf24b}
&\sum_{k=0}^{k=t} \delta^{t-k} \nabla f_k(\beta^{*(k)}_{inc})^{T} \left(\beta^{*(0)} - \beta^{*(t)}_{inc} \right) \nonumber \\&= - \sum_{j=0}^{j=t-1}\sum_{k=j+1}^{k=t} \delta^{t-j} \left(\beta^{*(0)} - \beta^{*(t)}_{inc} \right) ^{T} \nabla^{2}f_{j}(\beta^{*(j)}_{inc}) \left(\beta^{*(k)}_{inc} - \beta_{inc}^{*(k-1)}\right) \nonumber \\
&= - \sum_{j=0}^{j=t-1} \delta^{t-j} \left(\beta^{*(0)} - \beta^{*(t)}_{inc} \right) ^{T} \nabla^{2}f_{j}(\beta^{*(j)}_{inc}) \left(\beta^{*(t)}_{inc} - \beta_{inc}^{*(j)}\right).
\end{alignat}

\noindent Lastly, combining (\ref{pf24b}) and other two terms in (\ref{pf22}) into (\ref{pf21}), and grouping terms by $\nabla^{2} f_{k}(\beta^{*(k)}_{inc})$, we get
\begin{alignat}{1}
\label{pf25}
& F_{t}(\beta^{*(0)}) - F_{t} (\beta^{*(t)}_{inc}) \nonumber \\
&= \sum_{k=0}^{k=t} \tfrac{\delta^{t-k}}{2} \left[ \|\beta^{*(0)}-\beta^{*(k)}_{inc}\|^2_{\nabla^{2} f_{k} (\beta^{*(k)}_{inc})} - \|\beta^{*(t)}_{inc}-\beta^{*(k)}_{inc}\|^2_{\nabla^{2} f_{k} (\beta^{*(k)}_{inc})} \right. \nonumber \\
& \ \ \ \ \ \ \ \ \ \ \ \ \ \ \ \ \ \ \ \left. -2\left(\beta^{*(0)} - \beta^{*(t)}_{inc} \right) ^{T} \nabla^{2}f_{k}(\beta^{*(k)}_{inc}) \left(\beta^{*(t)}_{inc} - \beta_{inc}^{*(k)}\right)\right] + {\mathcal O}(\bar{\gamma}^3) , \nonumber \\
&= \sum_{k=0}^{k=t} \tfrac{\delta^{t-k}}{2} \|\beta^{*(0)}-\beta^{*(t)}_{inc} \|^{2}_{\nabla^{2} f_{k}(\beta^{*(k)}_{inc})} + {\mathcal O}(\bar{\gamma}^3), \nonumber \\
& \geq \sum_{k=0}^{k=t}\tfrac{\delta^{t-k} m_k}{2} \|\beta^{*(0)}-\beta^{*(t)}_{inc} \|^{2} + {\mathcal O}(\bar{\gamma}^3).
\end{alignat}
where $m_k$'s are the respective strong convexity constants for $f_k$'s. Again, $\sum_{k=0}^{k=t} \delta^{t-k} m_k \geq \lambda > 0$ ensures strict improvement.
\end{proof}

\section{Results on Movie Lens}
\label{appendix:movie}

\begin{figure}[h]
  \centering
  \includegraphics[width=0.65\linewidth]{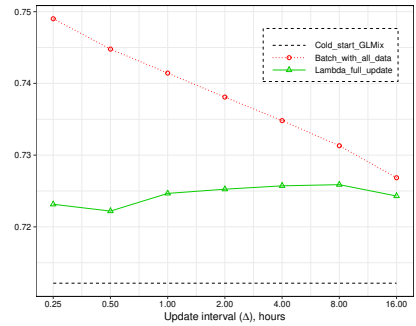}
  \vspace{-0.1in}
  \caption{Change in ROC-AUC, as we reduce the model update frequency. The point $t$, on the x-axis denotes the model being updated every $t$ hours - in the incremental case using ONLY data in the last $t$ hours and in the batch case using the entire history including the data in the last $t$ hours. $\delta=0.95$.}
  \label{movie-lens-batch}
  \vspace{+0.3in}
\end{figure}

We observe similar results on the publicly available Movie Lens dataset. Figure \ref{movie-lens-batch} shows the performance of the incremental update (LL) as compared to the theoretical upper bound (IBU) and the lower bound (NU).

\clearpage

\end{document}